%% file: main.tex
\documentclass[10pt,twocolumn,letterpaper]{article}

\usepackage{cvpr}              
\usepackage{subcaption}
\usepackage{booktabs}
\usepackage{multirow}
\usepackage{booktabs}
\usepackage{colortbl}
\usepackage{xcolor}
\usepackage{pifont}
\usepackage{float}
\usepackage[accsupp]{axessibility} 
\usepackage[pagebackref,breaklinks,colorlinks,allcolors=cvprblue]{hyperref} 
\definecolor{headerblue}{RGB}{221,235,247}
\definecolor{oursgreen}{RGB}{217,242,221}
\input{preamble}
\definecolor{cvprblue}{rgb}{0.21,0.49,0.74}
\usepackage[pagebackref,breaklinks,colorlinks,allcolors=cvprblue]{hyperref}
\usepackage{graphicx} 
\usepackage{svg}      
\def\paperID{10} 
\def\confName{CVPR}
\def\confYear{2026}

\title{GeoMeld: Toward Semantically Grounded Foundation Models for Remote Sensing}

\author{
Maram Hasan$^{1}$ \quad
Md Aminur Hossain$^{1, 2}$ \quad
Savitra Roy$^{1}$ \quad
Souparna Bhowmik$^{1}$ \quad
Ayush V. Patel$^{2}$  \\
Mainak Singha$^{3}$ \quad
Subhasis Chaudhuri$^{1}$ \quad
Muhammad Haris Khan$^{4}$ \quad
Biplab Banerjee$^{1}$\\
\\
$^{1}$Indian Institute of Technology Bombay \quad
$^{2}$Space Applications Centre, ISRO \\
$^{3}$University of Trento \quad $^{4}$Mohamed bin Zayed University of Artificial Intelligence\\
\\
}

\begin{document}
\maketitle

\input{sec/0_abstract}    
\input{sec/1_intro}

\input{sec/22_RW}
\input{sec/3_Dataset}

\input{sec/4_FM}
\input{sec/5_Experiments}
\input{sec/6_Conclusion}
{
    \small
    \bibliographystyle{ieeenat_fullname}
    \bibliography{main}
}


\end{document}

%% file: sec/0_abstract.tex
\begin{abstract}

Effective foundation modeling in remote sensing requires spatially aligned heterogeneous modalities coupled with semantically grounded supervision, yet such resources remain limited at scale. We present GeoMeld, a large-scale multimodal dataset with approximately 2.5 million spatially aligned samples. The dataset spans diverse modalities and resolutions and is constructed under a unified alignment protocol for modality-aware representation learning. GeoMeld provides semantically grounded language supervision through an agentic captioning framework that synthesizes and verifies annotations from spectral signals, terrain statistics, and structured geographic metadata, encoding measurable cross-modality relationships within textual descriptions. To leverage this dataset, we introduce GeoMeld-FM, a pretraining framework that combines multi-pretext masked autoencoding over aligned modalities, JEPA representation learning, and caption–vision contrastive alignment. This joint objective enables the learned representation space to capture both reliable cross-sensor physical consistency and grounded semantics. Experiments demonstrate consistent gains in downstream transfer and cross-sensor robustness. Together, GeoMeld and GeoMeld-FM establish a scalable reference framework for semantically grounded multi-modal foundation modeling in remote \linebreak sensing.~\href{https://github.com/MaramAI/GeoMeld}{https://github.com/MaramAI/GeoMeld}



\end{abstract}

%% file: sec/1_intro.tex
\section{Introduction}

Remote Sensing (RS) foundation models have advanced rapidly in recent years, driven by large-scale self-supervised learning and increasing availability of Earth observation imagery~\cite{huang2025survey}. RS foundation models can learn generalizable, transferable features from diverse sensors, most commonly optical and SAR, through self-supervised and weakly supervised learning. Such models enable a wide range of downstream
applications including land use classification, object/change detection, road extraction, and visual question answering, while reducing annotation costs and ensuring scalability~\cite{swartzman2025prithvi}.

Despite this progress, today’s pretraining resources remain fragmented across sensing modalities and supervision types. Most large-scale datasets are single-modality with text (e.g., optical~\cite{shabbir2025geopixel}, synthetic aperture radar (SAR)~\cite{ma2025sarchat} or optical–SAR combinations~\cite{shu2025earthmind}) and provide limited semantic annotations such as scene labels or land-cover categories. Although such supervision is effective for narrowly defined tasks, it does not fully capture the relational, environmental, and cross-modal structure inherent to geospatial data~\cite{nedungadi2024mmearth}. Furthermore, large-scale datasets with aligned natural-language captions are rare; many benchmarks contain fewer than one million samples~\cite{yuan2024chatearthnet}.

\begin{table*}[t]
\centering
\tiny
\renewcommand{\arraystretch}{1.12}

\begin{tabular}{p{1.5cm} p{3.7cm} p{1.6cm} p{1.3cm} p{1.6cm} c p{1.8cm}}
\toprule
\rowcolor{headerblue}
\textbf{Dataset} & \textbf{Modalities} & \textbf{Scale} & \textbf{Resolution} & \textbf{Tile Size} & \textbf{Meta-data} & \textbf{Foundation Model} \\
\midrule

MMEarth & S1, S2, Elevation, Slope, ETH-GCHM, Dynamic World, ESA WorldCover, Biome, Ecoregion, ERA5, Geo, Date & 1.2M locations & 10m & 128×128 & \ding{51} & MP-MAE \\

Skyscript & RGB + Text & 2.6M & 0.1m–30m & Variable & \ding{51} & SkyCLIP \\

Skysense & RGB, 10-band MS, SAR (VV/VH) & 21.5M & 0.3m, 10m & 2048×2048  64×64 & \ding{51} & SkySense \\

RS5M & RGB, S1, S2 (12-band), MS, Text & ~5M  & 0.5m–30m & Variable & \ding{51} & GeoRSCLIP \\

ChatEarthNet & S2, ESA WC 2020, Text & 173k  & 10m & 256×256 & \ding{51} & -- \\




Satlas Pretrain & S2, NAIP & 856k  & 1m / 10m & 8192×8192, 512×512 & \ding{51} & SatlasNet \\


BigEarthNet-MM & S1, S2 & 590k  & 10–60m & 120×120 to 20×20 & \ding{51} & -- \\

GRAFT & RGB, Flickr Images & 18.9M & 1m / 10m & 224×224 & \ding{51} & -- \\

SkySense V2 & HR RGB, MS (S2), SAR (S1), Text & 21M & High-res & 2048×2048 & \ding{51} & SkySense V2 \\

TerraFM & S1, S2 & 18.7M & 10–60m & 10.68km×10.68km & \ding{51} & TerraFM-B/L \\

SSL4EO & S1, S2 & 251k locations & 10m & 264 × 264 pixels & \ding{51} & CROMA \\

Prithvi-EO-2.0 & Harmonized S2 and Landsat & 4.2M & 30m & 256×256 & \ding{51} & Prithvi-EO-2.0 \\

GeoPlex & VHR, SPOT, S1, S2, Landsat, MODIS, ALOS-2 & ~1.95M & 0.2m–250m & 60m–1280m & \ding{55} & AnySat \\

\midrule
\rowcolor{oursgreen}
\textbf{GeoMeld (ours)} & S2,
S1(VV,VH,HH,HV)
ESA World Cover,
Dynamic World,
ASTER-DEM,
Canopy Height, location & 2.5M & 1m,10m & 128x128,
1280x1280 & \checkmark & GeoMeld-FM \\

\bottomrule

\end{tabular}
\caption{Comparison of major multi-modal and vision-language Earth Observation datasets.}
\label{tab:dataset_comparison}
\end{table*}

Recent RS foundation models have explored self-supervised and predictive learning at scale~\cite{wang2023ssl4eo}. However, most models operate within vision-only paradigms~\cite{wang2024skyscript} and rely primarily on spectral-band objectives, without incorporating structured language supervision~\cite{danish2025terrafm}. Even when multiple sensing modalities are included, supervision is often organized around modality-specific reconstruction or sensor-level objectives~\cite{fuller2023croma, danish2025terrafm}, without an explicit semantic alignment mechanism that jointly ties heterogeneous physical signals and language within a joint representation space~\cite{zhang2024rs5m}. This fragmentation lacks a semantic anchor that complements physical cross-sensor alignment. In particular, the integration of language supervision with structured multi-modal inputs at scale remains underexplored~\cite{ guo2024skysense}.

To address these structural gaps, we introduce \textbf{GeoMeld}, a large-scale multi-modal dataset designed to support modality-aware and semantically grounded foundation models in remote sensing. GeoMeld contains approximately 2.5 million spatially aligned samples spanning Sentinel-2 optical imagery, high-resolution NAIP imagery, multi-polarization Sentinel-1 SAR, elevation (ASTER-DEM), canopy height, land-cover products (ESA WorldCover and Dynamic World), and geographic metadata. Each sample is constructed under a alignment protocol, forming a structured tuple that enables structured cross-modal learning. Furthermore, we present an agentic multi-modal captioning framework that generates captions grounded in multiple modalities, including spectral measurements, terrain statistics, water presence indicators, and external geographic tags. Verification stages ensure consistency between textual claims and measurable geospatial attributes. Building on GeoMeld, we present GeoMeld-FM, a pretraining framework that combines multi-pretext masked reconstruction (MP-MAE)~\cite{nedungadi2024mmearth}, JEPA-based predictive representation learning~\cite{fei2023jepa}, and caption–vision contrastive alignment in a multimodal setting.  We evaluate GeoMeld-FM on downstream tasks and transfer settings, and quantify the impact of each component via ablation studies. By integrating large-scale multimodal alignment with semantically grounded language supervision, our approach provides a scalable foundation for modality-aware representation learning in remote sensing.

%% file: sec/22_RW.tex
\section{Related Works}

Large-scale multimodal fusion has become central to remote sensing foundation models. Early benchmarks such as BigEarthNet-MM~\cite{sumbul2021bigearthnet} and CROMA~\cite{fuller2023croma} demonstrated the benefit of spatially aligned Sentinel-1 and Sentinel-2 data for supervised and self-supervised learning. TerraFM~\cite{danish2025terrafm}, SSL4EO~\cite{wang2023ssl4eo}, and Prithvi-EO-2.0~\cite{swartzman2025prithvi} extended large-scale pretraining through masked autoencoding, self-distillation, and predictive objectives over aligned multi-sensor grids and time series. MMEarth~\cite{nedungadi2024mmearth} further increased modality diversity by aligning optical, SAR, elevation, canopy height, and environmental variables at pixel level, while SkySense~\cite{guo2024skysense} explored multi-temporal cross-sensor learning within a teacher–student framework. Cross-resolution integration has also been studied in SatlasPretrain~\cite{bastani2023satlaspretrain}, AnySat~\cite{astruc2025anysat}, and GRAFT~\cite{mall2023remote}, which combine heterogeneous sensors across spatial scales  within a shared spatial region.

In parallel, language supervision has developed through caption-style and instruction-oriented datasets. RS5M~\cite{zhang2024rs5m}, SkyScript~\cite{wang2024skyscript} construct image–text corpora using web data, rule-based templates derived from OSM tags. While RemoteCLIP~\cite{liu2024remoteclip} converts segmentation and detection annotations into templated captions over RGB imagery. Instruction-tuned resources such as EarthDial-Instruct~\cite{soni2025earthdial}, GeoChat-Instruct~\cite{kuckreja2024geochat}, SARChat-Bench2M~\cite{ma2025sarchat}, RSVL3M~\cite{hu2025ringmo}, and EarthGPT~\cite{zhang2024earthgpt} reformulate detection and classification annotations into conversational supervision. Region- and pixel-level grounding has been explored in EarthMarker~\cite{zhang2024earthmarker}, GeoPixel~\cite{shabbir2025geopixel}, and SkySenseGPT~\cite{luo2024skysensegpt}. 

While these works advance multimodal fusion or language alignment, they typically focus on either vision-only multi-sensor pretraining or image–text alignment derived from annotations. GeoMeld integrates spatially aligned heterogeneous modalities with semantically grounded captions derived from measurable geospatial signals and and trains them jointly with text supervision. (see Table~\ref{tab:dataset_comparison}.)



%% file: sec/3_Dataset.tex
\section{Dataset Construction}
Contemporary Earth observation archives show a clear geographic bias toward North America and Europe. To create a balanced pre-training dataset, we set a target extraction frame of about 2.5 million independent geographic points. We achieved this using three distinct sourcing strategies to ensure ecological diversity and spatial fairness.

\begin{figure*}[t]
    \centering
    \begin{subfigure}[t]{0.52\linewidth}
        \centering
\includegraphics[width=0.95\linewidth]{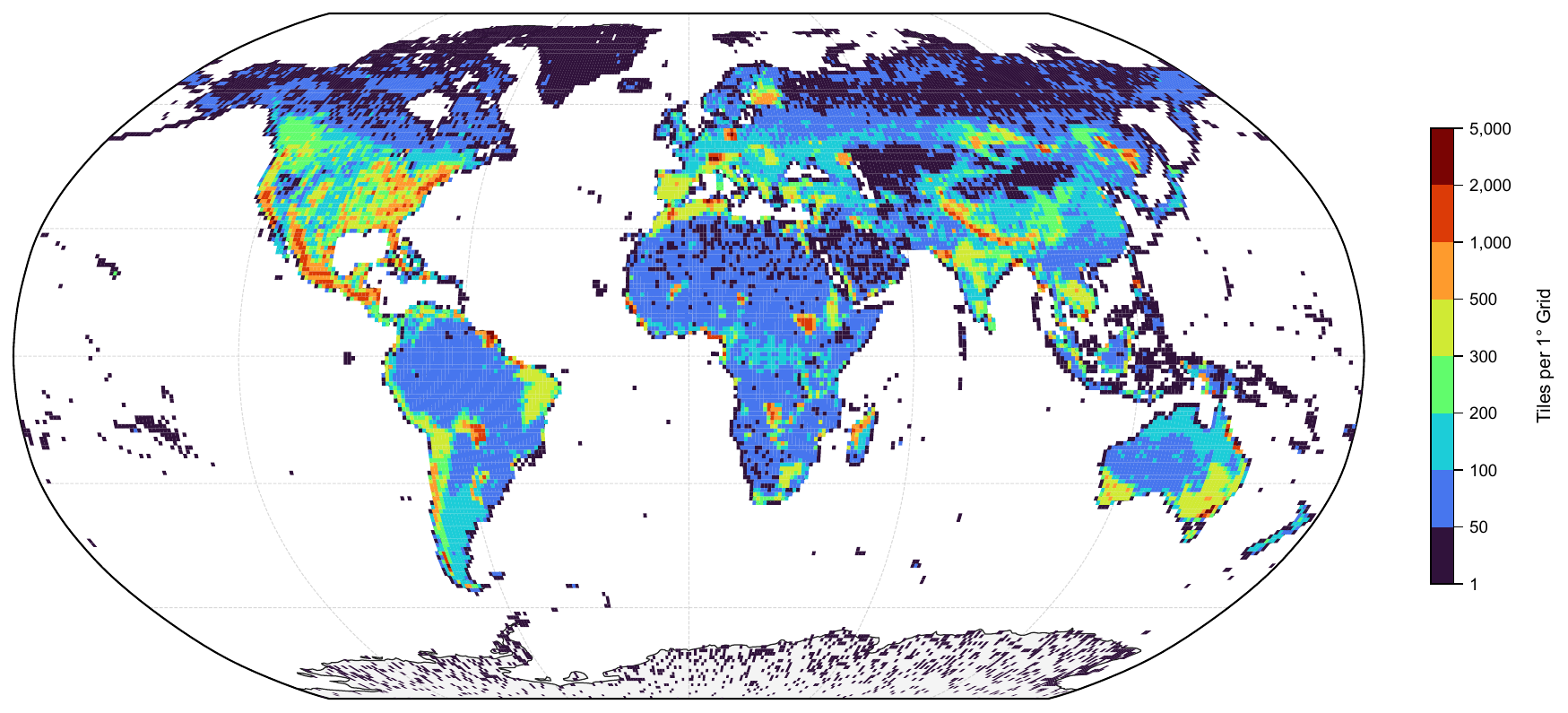}
        \caption{}
        \label{fig:short-a}
    \end{subfigure}
    \hfill
    \begin{subfigure}[t]{0.46\linewidth}
        \centering
       \includegraphics[width=0.95\linewidth]{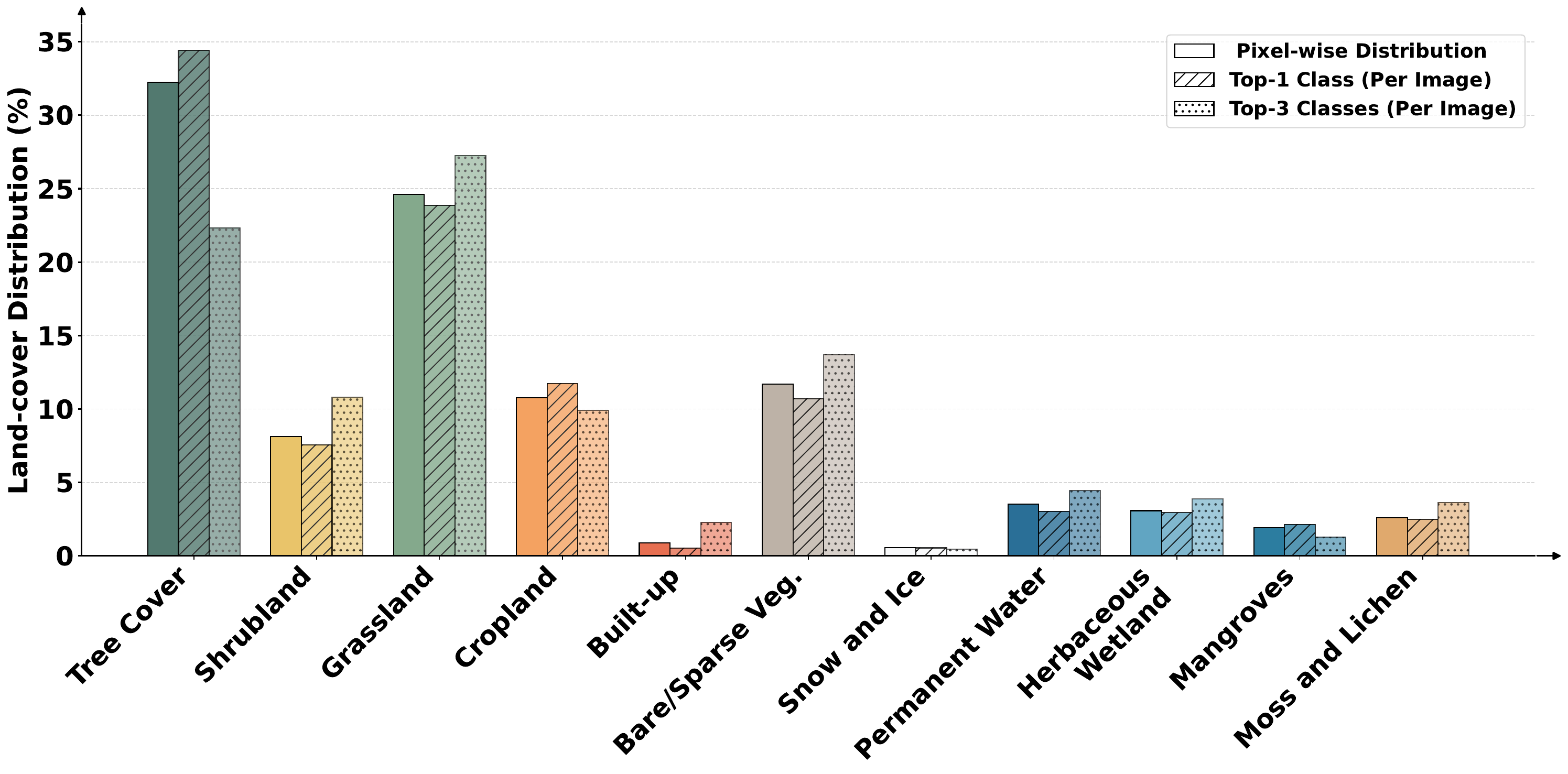}
       
        \caption{}
        \label{fig:short-b}
    \end{subfigure}
    \caption{ (a) Global spatial distribution of datapoints presented in 1\textdegree x1\textdegree cell, representing the number of tiles in 10.68 × 10.68 $km^2$. (b) Land-cover distribution in GeoMeld showing pixel-wise coverage, dominant (top-1) class, and top-3 classes per tile, highlighting broad biome diversity under landscape-scale sampling.}
    \label{fig:short}
\end{figure*}

\subsection{Geographic Sourcing  }

The geographic construction of GeoMeld is guided by three principles: biome balance, cross-dataset robustness, and global equity.  First, we adopt a biome-stratified sampling strategy to promote broad representation across major terrestrial ecosystems. By leveraging large-scale global repositories, we construct a geographically diverse coordinate pool spanning varied environmental regimes, reducing the dominance of common land-use types (e.g., agriculture or urban areas) while improving coverage of underrepresented biomes such as tundra, wetlands, and mangroves. This mitigates long-tailed geographic bias commonly observed in large remote sensing corpora.

Second, to enhance cross-dataset generalization and spatial robustness, we integrate spatial anchors derived from large-scale remote sensing datasets while preventing spatial overlap with their original imagery. Only coordinate geometries are retained as neutral extraction points, enabling multi-modal retrieval without inheriting annotations or captions. This strategy reduces data leakage and minimizes spatial autocorrelation between training sources.

Finally, we introduce additional custom geographic sampling focused on historically underrepresented regions across Africa, South America, and Asia. This targeted expansion addresses the geographic imbalance prevalent in many large-scale vision datasets, which disproportionately emphasize North American and European regions. By explicitly incorporating samples from the Global South, GeoMeld improves environmental diversity and supports more globally representative foundation modeling.

Together, these steps yield a geographically diverse, biome-aware, and spatially independent coordinate foundation for multimodal data extraction. Figures~\ref{fig:short-a} and~\ref{fig:short-b} illustrate the spatial and land-cover distributions of the dataset, respectively.

\subsection{Modality Alignment}

GeoMeld employs a spatially and temporally consistent multi-modal alignment protocol to ensure that heterogeneous sensing layers describe the same physical surface conditions. Each geographic coordinate serves as a fixed spatial anchor defining a standardized field of view, within which all modalities are retrieved and harmonized. Multi-resolution inputs, including optical, SAR, elevation, canopy structure, and land-cover products, are resampled to a common spatial grid to enable direct cross-modality correspondence at the pixel level.

Temporal alignment is handled through anchor-based retrieval to reduce cross-seasonal inconsistencies between sensors. Rather than opportunistically selecting acquisitions, each spatial anchor is associated with a controlled temporal reference that governs the retrieval window of all satellite-derived layers. This strategy ensures that modalities remain physically coherent while preserving natural seasonal variability across the dataset.  
\subsection{Agentic Caption Generation Framework}
\label{sec:agentic_framework}

\begin{figure*}

        \centering
\includegraphics[width=\linewidth, height=6.5cm]{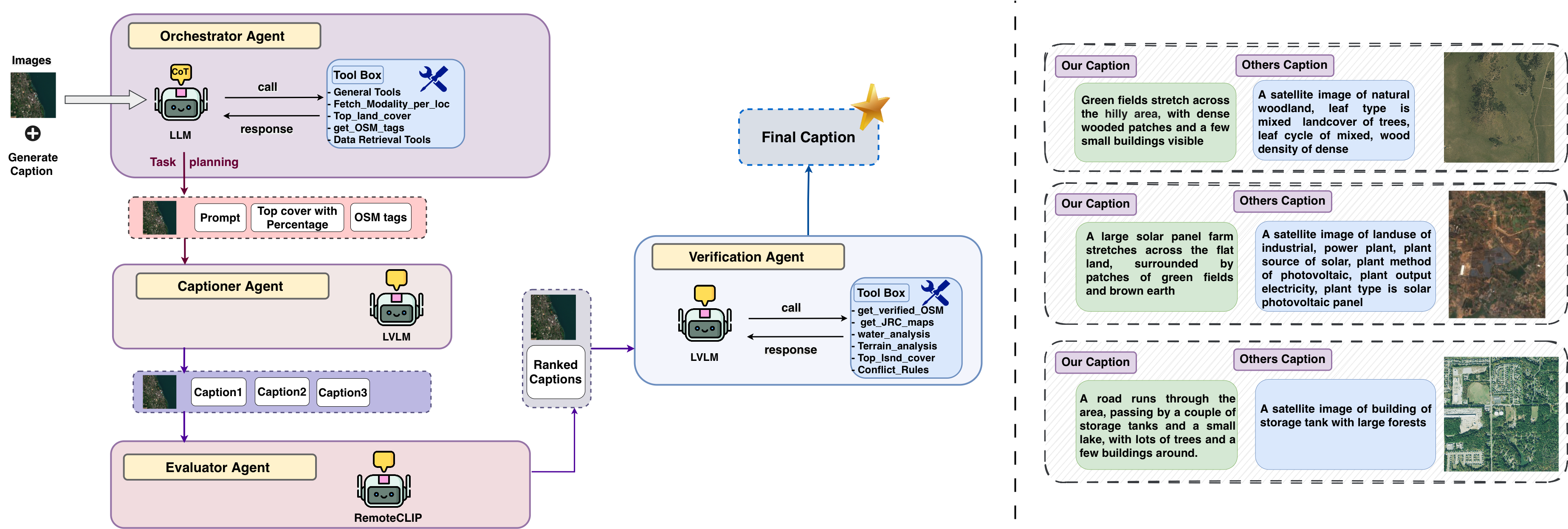}
        \caption{ Agentic framework for generating semantically grounded captions. An Orchestrator aggregates modality-specific signals and metadata, a Captioner produces multiple candidate descriptions, an Evaluator ranks them via vision–text alignment, and a Verification agent enforces consistency with structured geospatial attributes. Comparision of generated captions against captions of other dataset (Right). }
        \label{fig:caption}
    \end{figure*}

Central to our approach is an agentic multimodal captioning framework. Rather than generating captions solely from visual appearance, we employ a multi-agent system that generate semantically grounded captions by integrating heterogeneous geospatial signals. The framework, Figure~\ref{fig:caption}. decomposes caption generation into a sequence of coordinated agents, each responsible for a well-defined semantic function. Rather than relying on a single monolithic language model, the process is structured as a multi-stage pipeline that progressively synthesizes, evaluates, and verifies candidate annotations.

Given an input sample consisting of spatially aligned modalities (e.g., RGB imagery and auxiliary layers), the framework follows four main stages: (1) semantic signal extraction, (2) multi-candidate caption generation, (3) semantic ranking and refinement, and (4) final verification.

\noindent \textbf{Orchestrator Agent.} It initiates the process by performing task planning and signal preparation. It retrieves modality-specific information and structured metadata associated with the input sample, including dominant land-cover statistics and geographic tags. These signals are consolidated into a structured prompt that guides subsequent caption generation, ensuring that the resulting captions are grounded in both visual evidence and auxiliary environmental context.

\noindent \textbf{Captioner Agent.} Instead of producing a single description, the agent generates multiple candidate captions conditioned on the optical image and structured signals from the Orchestrator, capturing alternative semantic interpretations of the scene. These candidates incorporate visual patterns together with metadata such as dominant land cover or geographic attributes.

\noindent \textbf{Evaluator Agent.}
To avoid selecting captions based solely on language fluency, we introduce an Evaluator Agent that ranks candidate captions according to their alignment with the image content. This ranking stage provides a structured score table over candidate captions, enabling selection based on cross-modal consistency rather than generative confidence alone.

\noindent \textbf{Verification Agent.}
The final stage performs explicit refinement and semantic verification. Starting from the highest-ranked candidate, the agent cross-checks textual claims against external geospatial signals, land-cover maps, hydrological indicators, terrain statistics, and structured geographic tags. Conflict detection rules identify inconsistencies between the caption and measurable environmental attributes;  When discrepancies are detected, the caption is revised to ensure physical and semantic consistency.

By decomposing caption generation into coordinated agents, the proposed framework produces semantically grounded descriptions that encode cross-modality relationships while reducing hallucination and metadata inconsistency. This design enables scalable and physically informed language supervision for GeoMeld.



\subsection{Dataset Statistics}

In our dataset, the coordinate pool is instantiated from three sources. We incorporate 1.2M centroids from the MMEarth repository~\cite{nedungadi2024mmearth} spanning 14 terrestrial biomes. In addition, 699,540 geographic anchors are extracted from SkyScript~\cite{wang2024skyscript} after filtering the original 5M records to remove spatial overlap; only coordinate geometries are retained, and all associated imagery and text annotations are discarded. These anchors support heterogeneous satellite and aerial retrieval, including Sentinel-2, Landsat 8/9, and sub-meter collections. Finally, 666,000 additional coordinates are generated via controlled spatial sampling and integrated into the retrieval index. All coordinates are stored as neutral extraction points for multi-modal alignment.

Each spatial anchor defines a 1280\,m $\times$ 1280\,m region of interest centered at the coordinate. All satellite-derived modalities are harmonized to a common 10\,m ground sampling distance, producing 128 $\times$ 128 arrays with pixel-wise spatial correspondence. Temporal consistency is enforced through an anchor-based strategy. For United States samples, the acquisition date of the associated 1\,m NAIP orthophoto defines the temporal reference. For anchors derived from external datasets, a pseudo-random monthly reference is assigned. For custom-generated samples, a deterministic temporal anchor is constructed by sampling a year between 2018–2021 and a month uniformly at random, with the day fixed to the 15th. The sampling process is seeded using the unique tile identifier to ensure reproducibility.

\begin{figure}

        \centering
\includegraphics[width=\linewidth]{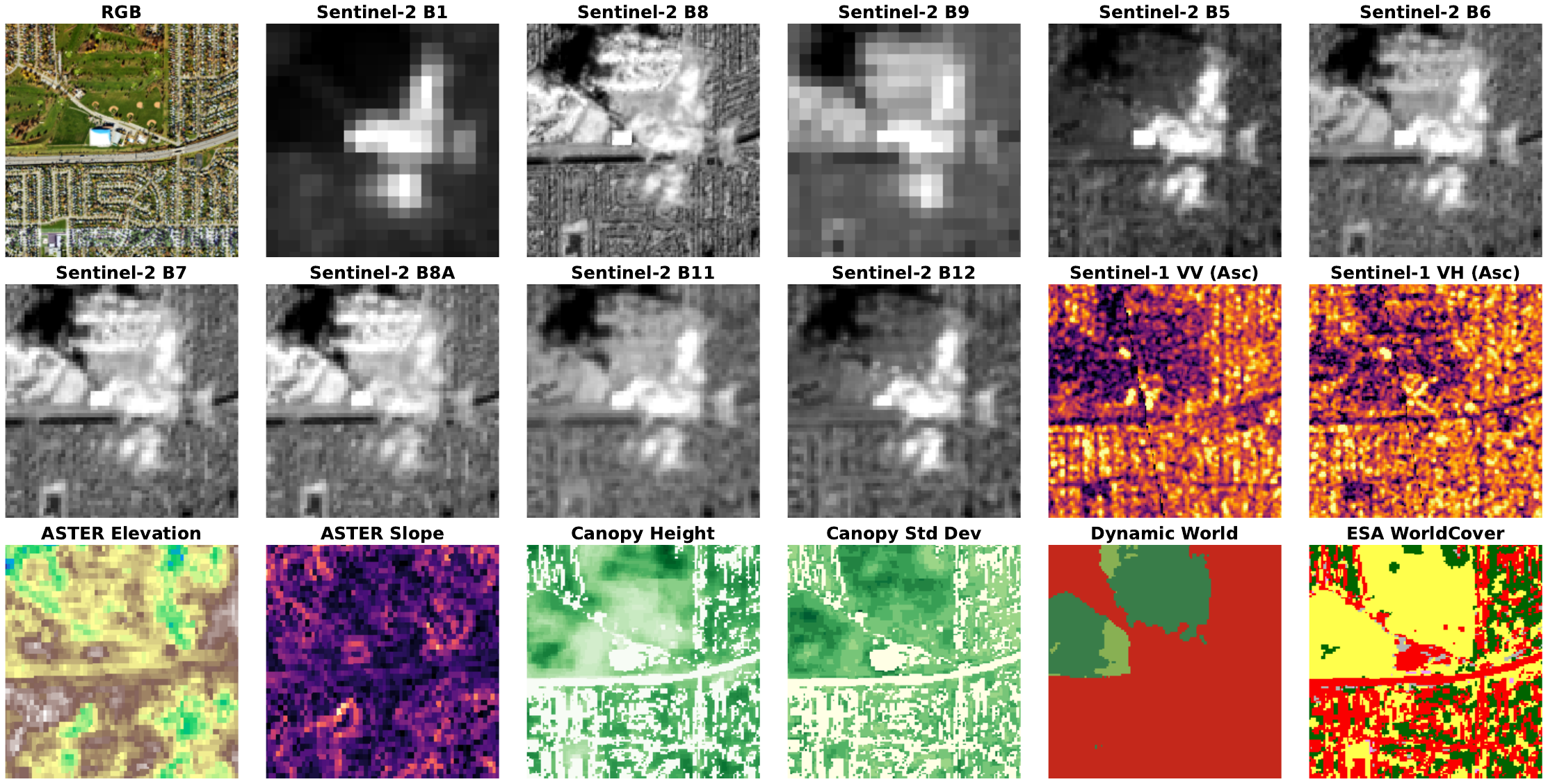}
        \caption{ A sample from GeoMeld showing spatially aligned multi-modal inputs derived through a unified alignment protocol.}
        \label{fig:modalities}
    \end{figure}

Given this temporal reference, Sentinel-2 (Level-2A) multi-spectral imagery is first retrieved under a strict cloud coverage constraint ($<10\%$). The selected Sentinel-2 timestamp then serves as the operational reference for dynamic modalities. Sentinel-1 SAR GRD backscatter (VV, VH, HH, HV; ascending and descending passes) and Dynamic World land-cover products are retrieved within a ±15-day window centered on the Sentinel-2 acquisition date. This hierarchical retrieval strategy ensures cross-sensor temporal coherence while accommodating realistic acquisition gaps across modalities. Figure~\ref{fig:modalities}. shows a multi-modal sample from GeoMeld of spatially aligned inputs.

%% file: sec/4_FM.tex
\section{Foundation Model Pretraining}

\subsection{Overview}
To provide baseline foundation-model results on GeoMeld, we introduce \textbf{GeoMeld-FM}, a pretraining framework that combines (i) \emph{multi-pretext masked autoencoding}~\cite{nedungadi2024mmearth} over aligned remote-sensing modalities and (ii) \emph{JEPA-style representation learning}~\cite{fei2023jepa} for Sentinel-2, together with (iii) \emph{caption--vision contrastive alignment} for semantically grounded supervision. The goal is to learn a representation space that captures (a) cross-sensor physical consistency (e.g., optical--SAR--terrain relationships) and (b) grounded semantics induced by the dataset's captions.

Each training sample consists of spatially aligned 10\,m grids ($128 \times 128$ arrays) for a subset of modalities including Sentinel-2 multispectral imagery, Sentinel-1 backscatter, ASTER-derived elevation, canopy height, and land-cover products (Dynamic World and ESA WorldCover), paired with a semantically grounded caption.

\subsection{Architecture}

\paragraph{Vision backbone and masking.}
We adopt the ConvNeXtV2~\cite{woo2023convnext} masked-autoencoder backbone as the vision encoder. During training, we apply patch masking to Sentinel-2 (12 bands) and feed only visible patches to the encoder. Let $x^{S2}\in\mathbb{R}^{12\times H\times W}$ be the S2 tensor (with $H=W=128$ after harmonization). A random mask $M$ selects the visible patch set $x^{S2}_{\mathrm{vis}}$. The encoder produces a latent sequence:
\begin{equation}
Z = E_{\theta}\!\left(x^{S2}_{\mathrm{vis}}\right)\in\mathbb{R}^{N_{\mathrm{vis}}\times d}.
\end{equation}
As in MAE~\cite{he2022masked}, masked patches are not embedded by the encoder; mask tokens are introduced only inside decoders.

\paragraph{Multi-pretext decoders (MP-MAE~\cite{nedungadi2024mmearth} over modalities).}
To exploit GeoMeld's aligned modalities, we attach lightweight modality-specific decoders that take the shared encoder latent $Z$ and predict each modality as a pretext task. Concretely, we use a decoder $D_m$ for each target modality
$m\in\{S2,S1,DEM,CH,DW,ESA\}$, where:
\begin{itemize}
    \item \textbf{Continuous rasters} ($S2$, $S1$, $DEM$, canopy height) are trained with masked reconstruction over patches.
    \item \textbf{Categorical land-cover products} (Dynamic World, ESA WorldCover) are trained as patch-wise (or pixel-wise) classification maps.
\end{itemize}
Let $\hat{x}^{m}$ denote the predicted output. For reconstruction decoders, we follow MAE and provide mask tokens for missing positions so the decoder predicts masked patches. For classification decoders, we predict a distribution over classes at each spatial location. This design encourages the encoder to learn latent features that are simultaneously useful for reconstructing optical reflectance, SAR backscatter, terrain geometry, vegetation structure, and land-cover semantics from the same underlying S2-driven representation.

    \begin{figure}[t!]

        \centering
\includegraphics[width=0.85\linewidth]{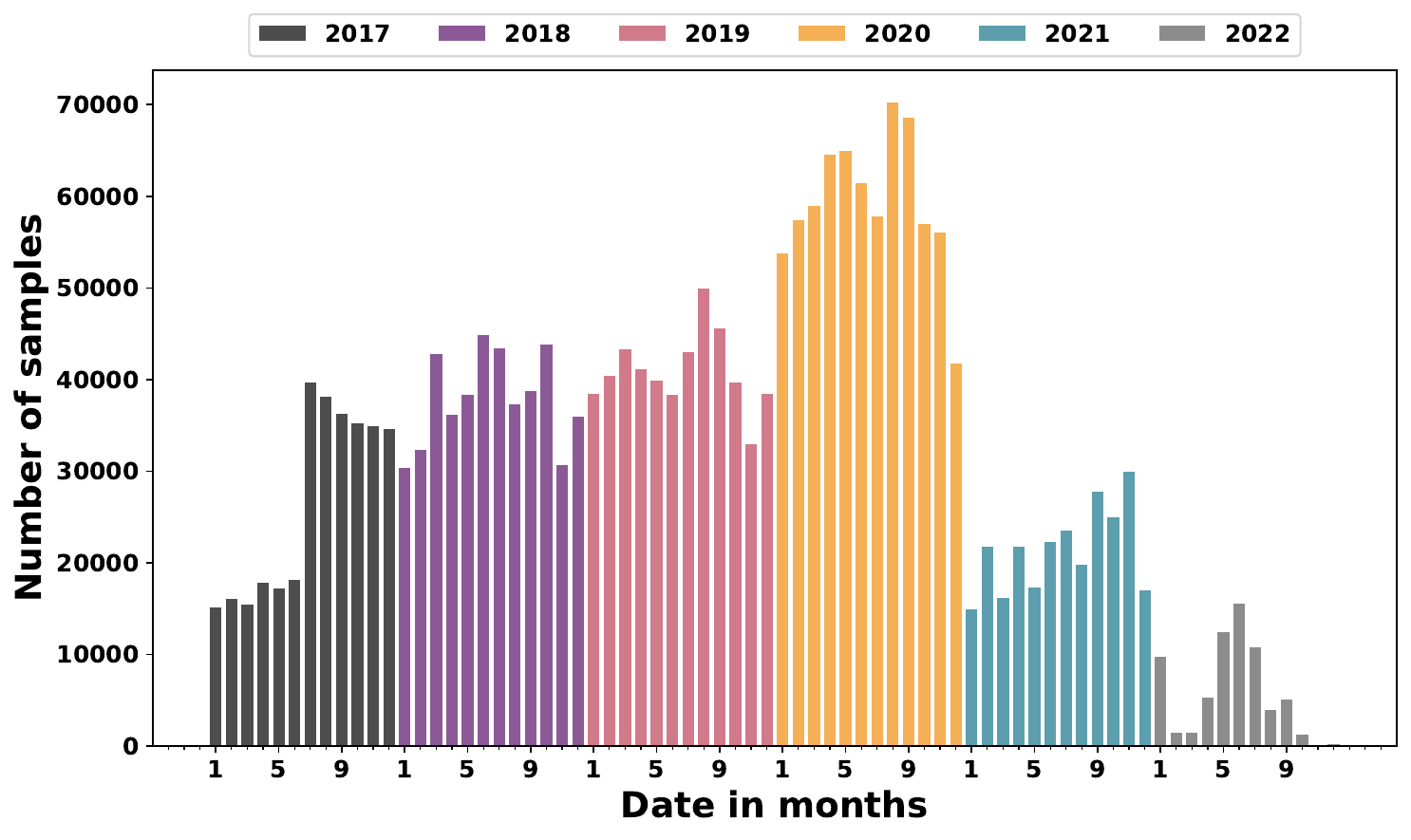}
        \caption{ Temporal distribution of GeoMeld samples, showing the number of samples per month and overall seasonal coverage}
        \label{fig:temporal}
    \end{figure}

\begin{figure*}[t]
    \centering
    \includegraphics[width=0.95\linewidth]{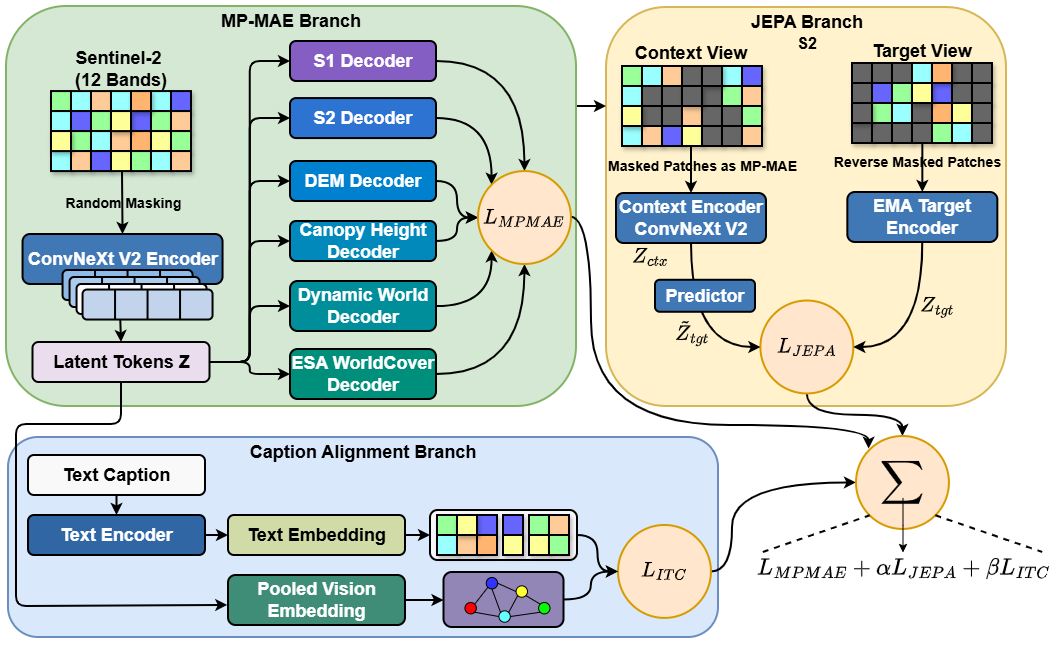} 
    \caption{\textbf{GeoMeld-FM pretraining architecture.}
    Sentinel-2 (12-band) tiles are patchified and randomly masked, and visible patches are encoded by a ConvNeXtV2 MAE encoder to produce latent tokens. The same masking pattern is used for the JEPA context encoder, while a separate non-overlapping mask forms the JEPA target view. Lightweight modality-specific decoders (MP-MAE) reconstruct or predict aligned modalities, including Sentinel-2, Sentinel-1, DEM, canopy height, and land-cover. In parallel, a JEPA branch forms context and target views of the same Sentinel-2 tile; a predictor maps the context to the target representation from an EMA teacher. For language grounding, captions are encoded and aligned with pooled vision embeddings using a symmetric contrastive (InfoNCE) objective. The model is trained jointly with three losses: MP-MAE reconstruction, S2-JEPA predictive loss, and caption-vision contrastive alignment.}

    \label{fig:mosaic_fm_arch}
\end{figure*}

\paragraph{JEPA branch for Sentinel-2 (predictive representation learning).}
Pixel reconstruction can over-emphasize low-level details. To additionally enforce high-level predictive structure, we integrate a JEPA objective on $S2$. We create two masked views of the same S2 tile: a \emph{context view} $x^{S2}_{\mathrm{ctx}}$ with visible patches under mask $M_{\mathrm{ctx}}$, and a \emph{target view} $x^{S2}_{\mathrm{tgt}}$ with visible patches under a different mask $M_{\mathrm{tgt}}$. The context encoder (shared with MAE) produces
\begin{equation}
Z_{\mathrm{ctx}} = E_{\theta}\!\left(x^{S2}_{\mathrm{ctx}}\right).
\end{equation}
A target encoder $E_{\xi}$ (updated by exponential moving average of $\theta$) produces
\begin{equation}
Z_{\mathrm{tgt}} = E_{\xi}\!\left(x^{S2}_{\mathrm{tgt}}\right).
\end{equation}
A predictor $P_{\phi}$ maps the context representation into the target latent space:
\begin{equation}
\hat{Z}_{\mathrm{tgt}} = P_{\phi}\!\left(Z_{\mathrm{ctx}}\right).
\end{equation}
This trains the model to predict latent representations of withheld spatial content from visible context rather than reconstructing pixels, improving semantic abstraction and robustness.

\paragraph{Caption encoder and vision--language alignment.}
GeoMeld provides a caption $c$ for each tile generated by our agentic captioning framework (Sec.~3.3). We encode the caption using a Transformer text encoder $T_{\psi}$ to obtain a text embedding:
\begin{equation}
t = T_{\psi}(c)\in\mathbb{R}^{d_t}.
\end{equation}
To align with text, we derive a single global tile embedding from the context latent sequence $Z_{\mathrm{ctx}}$ using pooling:
\begin{equation}
v = \mathrm{Pool}\!\left(Z_{\mathrm{ctx}}\right)\in\mathbb{R}^{d}.
\end{equation}
We then apply learnable projection heads $g_v(\cdot)$ and $g_t(\cdot)$ to map $v$ and $t$ into a shared contrastive space:
\begin{equation}
v' = g_v(v),\qquad t' = g_t(t).
\end{equation}
This alignment grounds the learned visual representation in semantically verified language signals, enabling caption--tile retrieval evaluation and improving transfer performance.

\subsection{Training Objectives}
Our final pretraining objective is a weighted sum of three losses:
\begin{equation}
\mathcal{L} = \mathcal{L}_{\mathrm{MPMAE}} + \alpha\,\mathcal{L}_{\mathrm{JEPA}} + \beta\,\mathcal{L}_{\mathrm{ITC}} .
\end{equation}

\vspace{-1em}

\paragraph{(1) Multi-pretext masked autoencoding loss $\mathcal{L}_{\mathrm{MPMAE}}$.}
For continuous modalities $m\in\{S2,S1,DEM,CH\}$, we reconstruct masked patches using an $\ell_1$ loss on masked positions $\Omega_m$:
\begin{equation}
\mathcal{L}_{\mathrm{rec}}^{m} =
\frac{1}{|\Omega_m|}
\sum_{i\in\Omega_m}
\left\|
\hat{x}^{m}_i - x^{m}_i
\right\|_1 .
\end{equation}
For categorical land-cover modalities $m\in\{DW,ESA\}$, we use cross-entropy over classes:
\begin{equation}
\mathcal{L}_{\mathrm{ce}}^{m} =
\frac{1}{|\Omega_m|}
\sum_{i\in\Omega_m}
\mathrm{CE}\!\left(\hat{y}^{m}_i, y^{m}_i\right) .
\end{equation}
The multi-task objective combines them with modality weights:
\begin{equation}
\mathcal{L}_{\mathrm{MPMAE}} = 
\sum_{\substack{m \in \{S2, S1, \\ DEM, CH\}}} \lambda_m \mathcal{L}_{\mathrm{rec}}^{m} 
+ \sum_{\substack{m \in \{DW, \\ ESA\}}} \lambda_m \mathcal{L}_{\mathrm{rec}}^{m}
\end{equation}

This component leverages the pixel-wise alignment of GeoMeld modalities (Figure~\ref{fig:mosaic_fm_arch}.) to induce cross-modality structure from a shared encoder.
\vspace{-1em}
\paragraph{(2) JEPA prediction loss $\mathcal{L}_{\mathrm{JEPA}}$ (S2 only).}
We supervise the predictor to match the target latent representation while stopping gradients through the target branch:
\begin{equation}
\mathcal{L}_{\mathrm{JEPA}} =
\frac{1}{|\Omega|}
\sum_{j\in\Omega}
\left\|
\hat{Z}_{\mathrm{tgt},j}
-
\mathrm{sg}\!\left(Z_{\mathrm{tgt},j}\right)
\right\|_2^2 ,
\end{equation}
where $\Omega$ indexes target tokens and $\mathrm{sg}(\cdot)$ denotes stop-gradient. The target encoder is updated via EMA:
\begin{equation}
\xi \leftarrow \tau\,\xi + (1-\tau)\,\theta .
\end{equation}
This objective promotes high-level predictive consistency under masking, complementing reconstruction-based pretext tasks.
\vspace{-1em}
\paragraph{(3) Image–Text Contrastive Loss $\mathcal{L}_{\mathrm{ITC}}$.}
We align tile embeddings and caption embeddings with a symmetric InfoNCE objective. For a batch of size $B$, define similarities:
\begin{equation}
s_{ij} = \frac{\left\langle v'_i, t'_j\right\rangle}{\tau_c} .
\end{equation}
The loss is:
\begin{equation}
\mathcal{L}_{\mathrm{ITC}}=
\frac{1}{2}\left(
\mathrm{CE}(s, \mathrm{diag})
+
\mathrm{CE}(s^\top, \mathrm{diag})
\right) ,
\end{equation}
encouraging matched tile--caption pairs to have higher similarity than mismatched pairs. This provides a direct mechanism for semantically grounded language supervision.

\subsection{Optimization and Practical Notes}
We train all components end-to-end, updating the vision encoder parameters $\theta$, modality decoders, the JEPA predictor $\phi$, and the text encoder $\psi$. The EMA target encoder parameters $\xi$ are updated only via the EMA rule above. In practice, we keep $\alpha$ moderate to avoid conflicts between pixel reconstruction (MP-MAE) and representation prediction (JEPA), and tune $\beta$ to control the strength of language grounding.


%% file: sec/5_Experiments.tex
\section{Experiments}

\subsection{Experimental Setup}
\vspace{-1em}
\noindent\paragraph{ Caption Generation.}
Our framework is implemented in Python using LangGraph for pipeline orchestration. Caption generation and refinement use the InternVL2.5-78B vision–language model, while candidate captions are ranked with RemoteCLIP-ViT-L-14. Land-cover statistics are obtained directly from the ESA WorldCover. 
\emph{Cross-modal Caption Grounding.} we used OpenStreetMap (OSM) tags to provide geographic context at three spatial levels: center-point features, surrounding objects, and area-level descriptors. Water presence is analyzed using a consensus of Dynamic World predictions with the JRC Global Surface Water dataset. Terrain context is derived through geomorphon-based classification~\cite{jasiewicz2013geomorphons} on ASTER-DEM.

\vspace{-1em}
\paragraph{Pretraining details.}
We pretrain GeoMeld-FM on the GeoMeld dataset using spatially aligned $128\times128$ tiles at 10\,m resolution. Sentinel-2 (12 bands) is used as the primary encoder input. We employ a ConvNeXtV2 backbone with patch masking ratio of 70\%. The model is trained for 150 epochs using AdamW with a learning rate of $1\times10^{-4}$, weight decay of 0.05, and cosine decay scheduling.  The JEPA target encoder is updated using exponential moving average (EMA) with momentum $\tau=0.996$. The loss weights are set to $\alpha=0.5$ for $\mathcal{L}_{JEPA}$ and $\beta=0.4$ for $\mathcal{L}_{ITC}$ unless otherwise stated. Effective batch size is 4096 across 4 GPUs (NVIDIA A100 80GB).

\begin{table*}[t]
\centering
\small
\setlength{\tabcolsep}{4pt}
\caption{Downstream evaluation results on GeoBench dataset. FT = full fine-tuning, LP = linear probing.}
\begin{tabular}{lcccc}
\hline
\textbf{Pretrain data} & \textbf{BigEarthNet20k (F1$\uparrow$)} \textbf{FT/LP} & \textbf{So2Sat20k (Acc.$\uparrow$)} \textbf{FT/LP} & \textbf{Cashew1k (IoU$\uparrow$)} \textbf{FT} & \textbf{SAcrop3k (IoU$\uparrow$)} \textbf{FT} \\
\hline
ImageNet & 55.7 / 25.9 & 36.6 / 24.0 & 77.1 & 26.7 \\
MMEarth (MP-MAE) & 67.1 / 43.3 & 54.6 / 43.8 & 79.8 & 38.2 \\
GeoMeld-S2 & 66.8 / 38.3 & 50.8 / 35.6 & 80.8 & 37.5 \\
GeoMeld-FM (Full) & \textbf{71.8 / 49.6} & \textbf{59.8 / 50.2} & \textbf{83.2} & \textbf{42.7} \\
\hline
\end{tabular}
\label{tab:geobench_mmearth_style}
\end{table*}

\begin{table*}[t]
\centering
\small
\setlength{\tabcolsep}{4pt}
\caption{Ablation study of GeoMeld-FM. MP = cross-modal multi-pretext reconstruction, LP = linear probing. Retrieval is reported as Recall@5 (R@5) for both image-to-text (I$\rightarrow$T) and text-to-image (T$\rightarrow$I) retrieval.}
\begin{tabular}{lccc|cccc|cc}
\hline
\textbf{Variant} & \textbf{MP} & \textbf{JEPA} & \textbf{ITC} & \textbf{BigEarth20k} & \textbf{So2Sat20k} & \textbf{Cashew1k} & \textbf{SAcrop3k} & \textbf{I$\rightarrow$T} & \textbf{T$\rightarrow$I} \\
 &  &  &  & \textbf{F1$\uparrow$ (LP)} & \textbf{Acc.$\uparrow$ (LP)} & \textbf{IoU$\uparrow$ (FT)} & \textbf{IoU$\uparrow$ (FT)} & \textbf{R@5$\uparrow$} & \textbf{R@5$\uparrow$} \\
\hline
S2-MAE baseline      & $\times$     & $\times$     & $\times$     & 38.3 & 35.6 & 80.8 & 37.5 & --   & --   \\
+ MP-MAE             & $\checkmark$ & $\times$     & $\times$     & 44.8 & 45.2 & 81.3 & 39.6 & --   & --   \\
+ MP-MAE + JEPA      & $\checkmark$ & $\checkmark$ & $\times$     & 46.1 & 46.9 & 82.4 & 40.8 & --   & --   \\
+ MP-MAE + ITC       & $\checkmark$ & $\times$     & $\checkmark$ & 45.4 & 45.8 & 81.9 & 40.1 & 31.4 & 33.8 \\
GeoMeld-FM (Full)    & $\checkmark$ & $\checkmark$ & $\checkmark$ & \textbf{49.6} & \textbf{50.2} & \textbf{83.2} & \textbf{42.7} & \textbf{37.8} & \textbf{39.6} \\
\hline
\end{tabular}
\label{tab:ablation}
\end{table*}


\noindent \textbf{Text encoder.}
Captions are encoded using a 6-layer Transformer text encoder initialized from scratch. The embedding dimension is set to 512. Projection heads for vision and text are implemented as 2-layer MLPs.

\subsection{Downstream Evaluation on GeoBench}

Following the evaluation protocol of MMEarth, we assess the quality of GeoMeld-FM representations on representative GeoBench~\cite{lacoste2023geo} downstream tasks spanning both image-level classification and pixel-level semantic segmentation. In particular, we adopt the same task family used in MMEarth to enable a direct and meaningful comparison with prior multi-modal pretraining work.

\noindent \textbf{Tasks.}
We evaluate on four representative GeoBench benchmarks:
\begin{itemize}
    \item \textbf{BigEarthNet20k} (multi-label land-cover classification),
    \item \textbf{So2Sat20k} (multi-class urban land-use / local climate zone classification),
    \item \textbf{Cashew1k} (semantic segmentation), and
    \item \textbf{SAcrop3k} (crop-type semantic segmentation).
\end{itemize}

\noindent \textbf{Evaluation protocol.}
We report both \emph{linear probing} (LP), where the pretrained encoder is frozen and only a lightweight task head is trained, and \emph{full fine-tuning} (FT), where all model parameters are updated. For the classification tasks, we report both FT and LP results, with each downstream model trained for 50 epochs. For the semantic segmentation tasks, following the MMEarth-style training setup, we report FT performance using a U-Net decoder on top of the pretrained encoder, trained for 100 epochs.

Table~\ref{tab:geobench_mmearth_style} shows that GeoMeld-FM outperforms both optical-only pretraining and the MMEarth multi-pretext baseline across all evaluated downstream tasks. The gains are particularly pronounced under linear probing, suggesting that the proposed training objectives encourage more transferable and semantically structured representations. Improvements on Cashew1k and SAcrop3k further indicate that the learned encoder preserves spatially meaningful features useful for dense prediction .

\subsection{Ablation Study}
To analyze the contribution of each major component in GeoMeld-FM, we conduct an ablation study under the same MMEarth downstream evaluation protocol. Specifically, we examine the contribution of: (i) cross-modality multi-pretext reconstruction (MP-MAE), (ii) JEPA-based predictive representation learning, and (iii) caption--vision contrastive alignment (ITC). We additionally evaluate bidirectional cross-modal retrieval to assess the shared vision--language embedding space learned through caption grounding. In particular, we report Recall (R@5) for both image-to-text retrieval (I$\rightarrow$T) and text-to-image retrieval (T$\rightarrow$I). 

Table~\ref{tab:ablation} summarizes the effect of progressively adding the three main components of GeoMeld-FM. Introducing cross-modality multi-pretext reconstruction already provides a substantial gain over the S2-only MAE baseline, confirming the value of aligned auxiliary modalities during pretraining. Adding the JEPA branch further improves both classification and segmentation performance, indicating that predictive latent-space supervision promotes more semantically structured and spatially robust representations. Incorporating ITC enables language grounding and yields meaningful bidirectional retrieval performance for both image-to-text and text-to-image matching, while also improving downstream transfer compared to the optical-only baseline. The full GeoMeld-FM model shows improved overall performance on downstream classification and segmentation tasks, suggesting that multi-modal reconstruction, predictive representation learning, and caption grounding contribute complementary benefits.

%% file: sec/6_Conclusion.tex
\section{Conclusion}
We introduced \textbf{GeoMeld}, a large-scale multi-modal remote sensing dataset comprising spatially aligned optical, SAR, elevation, canopy height, and land-cover modalities paired with semantically grounded captions.  Unlike prior datasets, GeoMeld explicitly integrates structured cross-sensor information with language supervision, supporting both vision-only and vision-language foundation models. We further proposed \textbf{GeoMeld-FM}, a pretraining framework combining multi-pretext masked autoencoding, JEPA-based predictive representation learning, and caption-vision contrastive alignment. Experiments on GeoBench downstream tasks demonstrate that cross-modality reconstruction, predictive latent learning, and language grounding provide complementary signals that improve transfer performance across classification, segmentation, and retrieval tasks. These results validate GeoMeld as a valuable resource for multi-modal foundation model research in Earth observation, with a strong baseline for future work.

\section{Acknowledgment}
This work was supported by the IITB-HRTI CoE on Computer Vision, Multimodal AI, and Geo-Intelligence, and the ANRF project (CRG/2023/004389).



